\documentclass[10pt,twocolumn,letterpaper]{article}

\usepackage{cvpr}              % To produce the CAMERA-READY version
\usepackage{adjustbox}
\usepackage{color,soul}
\usepackage{balance}

\usepackage[dvipsnames]{xcolor}

\definecolor{cvprblue}{rgb}{0.21,0.49,0.74}
\usepackage[pagebackref,breaklinks,colorlinks,citecolor=cvprblue]{hyperref}
\usepackage{multirow}
\usepackage{multicol}
\usepackage{booktabs}

\def\paperID{8} % *** Enter the Paper ID here
\def\confName{CVPR}
\def\confYear{2024}

\title{Adversarial Identity Injection for Semantic Face Image Synthesis}

\author{Giuseppe Tarollo$^1$, Tomaso Fontanini$^1$, Claudio Ferrari$^1$, Guido Borghi$^2$, Andrea Prati$^1$\\
$^1$ Department of Engineering and Architecture, University of Parma, Parma, Italy\\
$^2$ Department of Computer Science and Engineering, University of Bologna, Cesena, Italy\\
{\tt\small \{claudio.ferrari2, tomaso.fontanini, andrea.prati\}@unipr.it}, {\tt\small guido.borghi@unibo.it}
}

\begin{document}
\maketitle

\begin{abstract}
Nowadays, deep learning models have reached incredible performance in the task of image generation.
Plenty of literature works address the task of face generation and editing, with human and automatic systems that struggle to distinguish what's real from generated.
Whereas most systems reached excellent visual generation quality, they still face difficulties in preserving the identity of the starting input subject. Among all the explored techniques, Semantic Image Synthesis (SIS) methods, whose goal is to generate an image conditioned on a semantic segmentation mask, are the most promising, even though preserving the perceived identity of the input subject is not their main concern. Therefore, in this paper, we investigate the problem of identity preservation in face image generation and present an SIS architecture that exploits a cross-attention mechanism to merge identity, style, and semantic features to generate faces whose identities are as similar as possible to the input ones. Experimental results reveal that the proposed method is not only suitable for preserving the identity but is also effective in the face recognition adversarial attack, \ie hiding a second identity in the generated faces.
\end{abstract}

\section{Introduction}\label{sec:introduction}
In recent years, deep learning models have reached outstanding results in image generation, with several architectures standing out in the field of human face generation~\cite{karras2019style,karras2020analyzing,rombach2022high} and editing~\cite{zhou2022codeformer,borghi2021automated,wang2021towards}. 
Indeed, through such powerful models, both humans and automated systems struggle to distinguish between real and generated face images~\cite{george2019biometric,scherhag2019face}. 
However, even the most realistic generative model has difficulty in preserving the perceived identity of the generated subject after reconstructing or manipulating a real face image of a specific individual~\cite{sinha2020identity,mori2012uncanny,di2024face}. Whereas this aspect is often neglected or partially discussed in related works, preserving the perceived identity is crucial to make synthetic data exploitable in biometrics applications.

Thus, in this paper, we investigate a possible solution to maximize the identity preservation property without sacrificing the generation quality, with a specific focus on face editing models. In this context, one of the most effective techniques to perform this task is through Semantic Image Synthesis (SIS)~\cite{zhan2023multimodal}. The goal of SIS is to generate realistic face images conditioning the model with a semantic mask, \ie an image in which each pixel represents a semantic class \eg hair, eyes, or mouth. Therefore, the semantic mask is a key element in defining the final shape of the edited face.

\begin{figure}[!t]
    \centering
    \includegraphics[width=\linewidth]{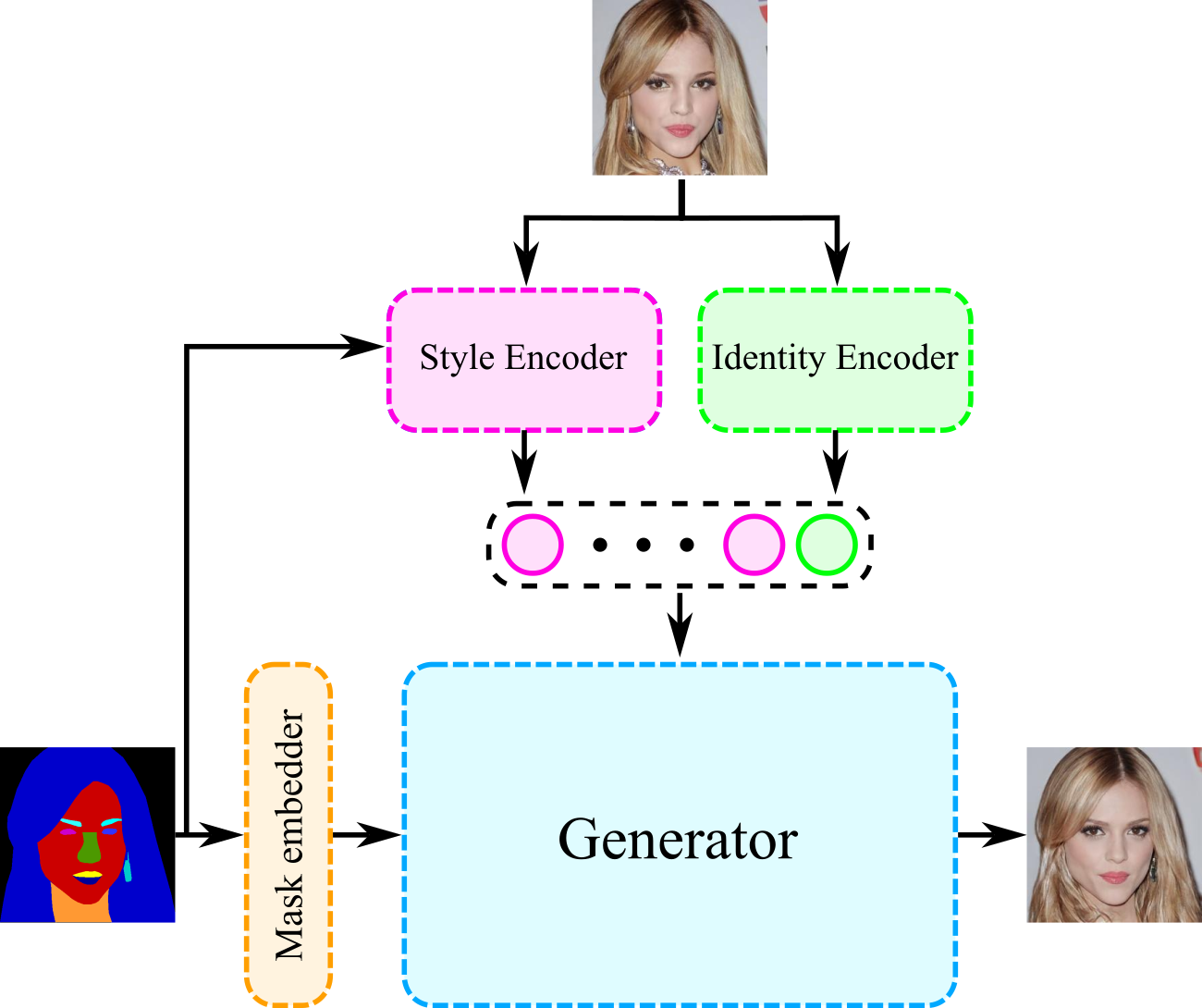}
    \caption{Overview of the proposed architecture. Starting from a face image, style and identity features are extracted through the encoders (Style Encoder ($\mathcal{E}_{s}$), Identity Encoder ($\mathcal{E}_{id}$) and Mask Embedder ($\mathcal{E}_{m}$)) and used by the Generator ($\mathcal{G}$), together with the semantic segmentation mask to generate the final image.}
    \label{fig:net}
\end{figure}

The clearest advantage of SIS methods is that the semantic mask can be used to learn explicit mappings between each semantic class and its style \ie texture. In doing so, modern SIS methods can independently generate, control, or manipulate the style of local face regions~\cite{tan2021diverse,zhu2020sean,fontanini2023semantic, fontanini2023automatic}. Whereas most SIS methods aim at learning such mapping to generate new images conditioned on the mask (\textit{noise-based})~\cite{park2019semantic, lee2020maskgan}, some literature methods such as~\cite{zhu2020sean,fontanini2023semantic} focus on extracting styles from a real (\textit{reference}) image in order to map them in the corresponding semantic regions (\textit{reference-based}). In this manner, it is specifically possible to perform editing of real images, for example by changing the hair color, makeup, and other different attributes in the generated samples. In this particular scenario, where the edited face belongs to a real individual, the ability to preserve the perceived identity is of utmost importance both if presented to a human observer or an automatic face recognition system~\cite{pini2021systematic}. 

Unfortunately, we observe that in semantic-based methods, especially for reference-based ones, the identity preservation of the edited face is not taken into account (\textit{e.g.}~\cite{tan2021diverse,zhu2020sean}). The large majority of available models neglect this aspect and grasp the input subject identity only to a limited extent, so lacking the ability to preserve it in the edited face as we show in our experimental evaluation (see Table~\ref{tab:comparison_rec}). As a consequence, state-of-the-art Face Recognition (FR) systems~\cite{schroff2015facenet, deng2019arcface} would struggle to match the identity of the reconstructed face with that in the input.

Therefore, in this paper, we propose a solution to inject the identity information into a reference-based semantic image synthesis architecture. In particular, the proposed architecture builds upon that proposed in~\cite{fontanini2023semantic}, and is composed of four different modules, as depicted in Figure~\ref{fig:net}: the Style Encoder and Identity Encoder models are responsible for extracting style and identity features, respectively, from the input face image. These features are then concatenated and fed as input to the Generator, responsible for the image generation. The Generator receives as input also the output of the Mask Embedder, which embeds the semantic information of the mask through a fully connected network; style, identity, and semantic information are finally merged through a cross-attention mechanism for face generation.

By exploiting the versatility of cross-attentions, we are able to condition the image generation with high-level information such as the identity, in addition to low-level style features, ultimately improving the identity similarity with respect to the input face. Nonetheless, another noticeable feature that arises with this design choice is the ability to change the identity embedding. In~\cite{fontanini2023semantic}, the model can be used to swap specific style embeddings, so to perform ``local'' style transfer, even if not explicitly trained to do so. Thus, we expect our design to let us change the identity embedding, so conditioning the generation of a face belonging to a subject $A$ with the identity of another subject $B$.  

Therefore, when injecting the identity embedding of the input face image, our model helps in preserving its perceived identity during the whole generation process. As a result, the generated face presents an identity that is closer to the input one, which can be appreciated both visually and quantitatively when it is presented to a face recognition model. Conversely, we can also use our method to swap the identity between two different subjects. In other words, we can concatenate the style embeddings of a subject $A$ with the identity embedding of another individual $B$. As a result, the generated face qualitatively looks like $A$, but it actually conceals the identity embedding of $B$. We wish to highlight here that, despite looking conceptually similar, this paradigm is different than common face-swapping methods~\cite{li2019faceshifter}. In fact, face swapping approaches aim at changing the perceived identity of a subject in a way such that a human observer easily recognizes the \textit{new} identity \ie the one that was swapped. Differently, our solution ``hides'' the identity in a way that a human observer can hardly tell the difference, but at the same time, it will fool recognition algorithms. In both tests, the proposed method achieves competitive results, overcoming the large part of literature methods. To summarize, the main contributions of this paper are listed in the following:
\begin{itemize}
    \item We explore the use of attention-based mechanism to merge identity, semantic and style information into the generation process of Semantic Image Synthesis (SIS);
    \item We test the proposed method in the task of identity preservation, and show our solution is promising in preserving the input identity during the generative process.
    \item We investigate the use of the proposed architecture in the task of adversarial attacks on face recognition, presenting an alternative pathway to achieve this goal.
\end{itemize}

\section{Related Work}
\noindent\textbf{Semantic Image Synthesis.} SIS models aim at generating images starting from a semantic mask. Several approaches were proposed to solve this task. 

Firstly, SPADE \cite{park2019semantic} proposed a spatially adaptive de-normalization module to modulate the activations with semantic information. Later, MaskGAN \cite{lee2020maskgan} proposed a method to manipulate human faces with semantic masks. MaskGAN was introduced simultaneously with SEAN \cite{zhu2020sean} which allowed to extract semantic styles from a reference images and apply them to the generated samples. Along this line, multiple methods were developed, like CLADE \cite{tan2021efficient} and INADE \cite{tan2021diverse}, which introduced the concept of instances in the semantic masks. Recently, Semantic-StyleGAN \cite{shi2022semanticstylegan} allowed to control the generation of StyleGAN images \cite{karras2019style} through semantic information. Finally, Semantic Diffusion \cite{wang2022semantic} adapted a diffusion model adding SPADE normalization layers in order to control the generation with semantic masks. Methods for semantic image synthesis are domain agnostic, meaning that they can be applied to several different scenarios \textit{e.g.} faces, outdoor scenes, objects. For this reason, domain-specific information such as the perceived identity as in the human face domain are neglected. Differently, the proposed model is specifically tailored for the human face domain.

\smallskip

\noindent\textbf{Adversarial attack on face recognition.}
Deep learning architectures have been proven susceptible to adversarial attacks~\cite{szegedy2013intriguing}, which are slightly-perturbed versions of
the original examples that eventually trick the networks into outputting a wrong prediction. The
peculiar characteristic of adversarial examples is that they are hardly distinguishable from their
``clean'' counterpart for the human eye, because they have been specifically optimized to have the
minimum possible perturbation. Adversarial attacks can be broadly divided into two categories: white-box or black-box. The former ones are the most difficult to counteract, as they assume full knowledge of the attacked system including model parameters, so that the attacker can exploit the model's gradient to craft the adversarial example. Black-box attacks are instead more difficult to craft since they only have access to the classifier prediction. Both can be also targeted or un-targeted. Targeted attacks aim at making the attacked model predict a \textit{specific class}, while un-targeted methods only care for making the model predict a wrong class. Despite being applicable whenever a classification task is involved, attacks can also be designed for specific domains and tasks, such as face recognition.
Attacks on face recognition can be divided in multiple categories. Firstly, gradient-based methods like \cite{goodfellow2014explaining, madry2017towards, dong2019efficient} aim at adding perturbations in the pixels, but suffer from common denoising models. Next, patch-based methods focus on printing on the images adversarial hat \cite{komkov2021advhat} or glasses \cite{sharif2016accessorize}, but in this case the attack is easily spottable. Finally, stealthy-based methods inject the adversarial attack in the face attributes \cite{yin2021adv, qiu2020semanticadv, jia2022adv}. On the other side, our system treats identity information as an additional style and therefore the generated samples will not be changed by adding glasses or makeup making the attack almost invisible. Recently, a new type of attack was proposed by Li \textit{et al.} \cite{li2023sibling} which utilizes an additional Attribute Recognition (AR) task to improve the attacking
transferability.

\section{Identity-conditioned Image Synthesis}\label{sec:ID_SIS}
The proposed system builds upon the very recent SIS model proposed by Fontanini \textit{et al}.~\cite{fontanini2023semantic} named $CA^2SIS$. We chose this specific architecture as, differently from the vast majority of SIS models that employ SPADE layers \textit{e.g.}~\cite{zhu2020sean,park2019semantic}, it uses spatial transformer blocks to condition the image generation with style features extracted from a RGB reference. 

The versatility of the cross-attention layers included in the spatial transformer blocks allowed us to design an alternative solution to inject identity information into the generator. This could not be done with standard SIS models based on SPADE as they require an explicit spatial mapping of the style features. While this mapping is straightforward when style features represent a well-defined class \textit{e.g.} hair, eyes, that becomes challenging if using features related to high-level concepts such as identity. In fact, there is no clear prior on such information; in other words, which face parts influence the the identity perception the most? To what extent? Whereas some literature works do provide some hints in this regard~\cite{ferrari2022makes, sinha2006face}, what contributes to recognizing an individual is actually a combination of facial features. This makes SPADE-like layers difficult to use. Cross-attentions instead provide a nice alternative since the spatial mapping is implicitly learned by the attention mechanism. This allows the model to learn how to optimally map the identity information into the generated face image without requiring prior intervention (Fig.~\ref{fig:att_map}). 

\subsection{Architecture}\label{subsec:architecture}
The objective of the original architecture is that of generating a photo-realistic image given a semantic segmentation mask and a reference image. It is composed by three modules: a Cross-Attention Generator $\mathcal{G}$, a Multi-Resolution Style Encoder $\mathcal{E}_{s}$, and a Mask Embedder $\mathcal{E}_{m}$. Necessary details are provided in the paragraphs below so to make the paper self-contained, but we refer the reader to~\cite{fontanini2023semantic} for a detailed description. 

\smallskip

\noindent
\textbf{Mask Embedder.} Let a semantic mask be a $C$-channel image $\mathcal{M} \in \mathbb{N}^{C \times H \times W}$, where each channel $\mathcal{M}_j$ is a binary image encoding the pixel-wise spatial location of a specific class \textit{e.g.} eyes, lips, hair. The module is an MLP that receives $\mathcal{M}$ and outputs $C$ embeddings of size $256$, one for each semantic class. These are reshaped to form $16 \times 16$ feature maps, and then stacked to form a mask descriptor $m_x = \mathcal{E}_{m}(\mathcal{M}) \in \mathbb{R}^{16 \times 16 \times C}$. The descriptor $m_x$ will be the input to the generator $\mathcal{G}$. 

\smallskip

\noindent
\textbf{Style Encoder.} The Style Encoder $\mathcal{E}_{s}$ extracts style features from the input RGB reference images $x$. Specifically, it is equipped with Grouped Convolutions, Group Normalization layers and skip connections, and is designed to extracts a style code $s_c \in \mathbb{R}^{256}$ for each semantic class $c$ by exploiting the mask $\mathcal{M}$. The style codes are concatenated to form a combined style code of size $256 \times C$, \textit{i.e.} $s_x = \mathcal{E}_{s}(x) \in \mathbb{R}^{1280}$.

\smallskip

\noindent
\textbf{Generator.} Finally, the generator $\mathcal{G}$ receives the mask descriptor as input and the style codes as condition, ultimately outputting a realistic image having the shape defined by the semantic mask, and the styles of the reference image, that is $\hat{x}=\mathcal{G}(m_x,s_x)$. More in detail, style codes are injected in the cross-attention ($CA$) layers of the Generator that are defined as follows:
\begin{equation}
     CA(Q,K,V) = \mathcal{S}\left(\frac{QK^T}{\sqrt{d}}\right)V
    \label{eq:att_map}
\end{equation}
\noindent where $Q = W_Q^{(i)} \cdot \phi^{(i)}$ is obtained from the projection of the flattened features $ \phi^{(i)}$ of previous convolutional layers, while $K = W_K^{(i)} \cdot \mathcal{E}_s(x_i)$ and $V = W_V^{(i)} \cdot \mathcal{E}_s(x_i)$ are computed from the style codes.
The Generator is also paired with a Discriminator $\mathcal{D}$ to exploit the adversarial loss.

\subsection{Identity Module}\label{subsec:identity_module}

Despite the original model itself can already preserve the perceived identity of the input face $x$ way better than other approaches, it does not allow for manipulating or changing it explicitly, although it is possible to swap style codes of a different face image $y$ for generating diverse images \textit{i.e.} $\hat{x}=\mathcal{G}(m_x,s_y)$. Given that the goal of this work is to explore an alternative pathway to preserve the identity of a subject $A$ or to conceal the identity of an individual $A$ into a face image of another subject $B$ without making the change being perceivable, we augmented this architecture by adding a pre-trained face recognition model $\mathcal{E}_{id}$. This module is used to extract an identity embedding from the input face $x$ \textit{i.e.} $id_x = \mathcal{E}_{id}(x)$, which is then used as additional style code for the generator (see Fig.~\ref{fig:net}). The idea is that in doing so, we can both increase the capability of the generator to preserve the original identity, while also being able to swap the identity code so to condition the generation with the identity embedding of a different individual. Ultimately, this leads the generated image to qualitatively appear as the original subject, at the same time concealing the identity information of a different subject. 

\begin{table*}[!th]
    \centering
    \begin{tabular}{c|c|c|c|c}
    \toprule
        \textbf{Method} & \textbf{IR152}  $\uparrow$ & \textbf{MobileFace}  $\uparrow$ & \textbf{FaceNet}  $\uparrow$ & \textbf{FID} $\downarrow$ \\
        \midrule
        
        SEAN \cite{zhu2020sean} & 0.51  & 0.73 & 0.74 & 18.7 \\
        V-INADE \cite{tan2021diverse}  & 0.23 & 0.49 & 0.43 & 18.3 \\
        Semantic StyleGAN \cite{shi2022semanticstylegan}   & 0.38 & 0.66 & 0.62 & 26.8\\
        \textbf{$CA^2SIS$}~\cite{fontanini2023semantic} & 0.52 & 0.75 & 0.74 & \textbf{15.8}\\
        \midrule
        \midrule
        \textbf{Ours} - FaceNet & \textbf{0.64} & \underline{0.80} & \textit{\textbf{0.90}}& 18.1\\
        \textbf{Ours} - ArcFace & \underline{\textit{0.62}} & \textbf{0.83} & \underline{0.81} & \underline{16.5} \\
        \bottomrule
    \end{tabular}
    \caption{Cosine similarity metric and FID comparison between original $x$ and reconstructed $\hat{x}$ faces when conditioning with its own identity embedding $id_x$. Injecting the identity information increases the similarity to a great extent for various FR models (see Sect.~\ref{subsec:id_preservation}). \textbf{Bold}=best result, \underline{underlined}=second best. 
    \textit{Italic} font indicates the validation FR architecture is the same as that used to train our model, but the pre-trained weights differ.}
    \label{tab:comparison_rec}
\end{table*}

\subsection{Identity Preservation Loss}\label{subsec:identity_loss}
In order to inject identity information into the Generator, we treat the identity embedding as an additional style code. In particular, we employed a pretrained face recognition model to extract an identity embedding from a reference image $x$ of some subject $i$. The embedding is then concatenated to the style codes extracted from $\mathcal{E}_{s}$, and the new identity-style representation is injected in the cross-attention layers of $\mathcal{G}$. Formally, we indicate this addition as $\hat{x}=\mathcal{G}(m_{x_i},s_{x_i},id_{x_i})$ During training, an \textit{identity preservation loss} $\mathcal{L}_{id}$ is employed in order to force the model to utilize this additional information. $\mathcal{L}_{id}$ is as follows:
\begin{equation}
    \mathcal{L}_{id} = 1 - cos\left(\mathcal{E}_{id}\left(\mathcal{G}\left(m_{x_i},s_{x_i},id_{x_i}\right)\right),\mathcal{E}_{id}(x_i)\right)
\end{equation}
\noindent where $\mathcal{G}(m_{x_i},s_{x_i},id_{x_i})$ is the generator output starting from reference mask $m_{x_i}$, style codes $s_{x_i}$ and identity embedding $id_{x_i}$. The term $cos$ is the cosine similarity function. This loss forces the generated samples to match the identity embeddings that are injected in the model during training. At inference time, the style codes and identity can be extracted from two difference images $x_i$ and $x_j$ resulting in a sample having the same appearance as $x_i$ but that will be recognized as $x_j$ by a FR network.

\subsection{Training Objective}
In addition to the identity preservation loss, during training, we employ a set of losses as in~\cite{fontanini2023semantic}. More in detail, we implemented an adversarial loss $\mathcal{L}_{adv}$, a feature matching loss $\mathcal{L}_{FM}$ \cite{wang2018high} and a perceptual loss $\mathcal{L}_{prc}$ \cite{johnson2016perceptual}. The full training objective becomes:

\begin{equation}
    \mathcal{L}_{tot} = \mathcal{L}_{adv} + \lambda_{FM}\mathcal{L}_{FM} + \lambda_{prc}\mathcal{L}_{prc} + \lambda_{id}\mathcal{L}_{id}
\end{equation}

\noindent where $\lambda_{FM}$, $\lambda_{prc}$ and $\lambda_{id}$ are the weights for feature matching, perceptual and identity preservation loss, respectively. More in detail, they are set during training as follows: $\lambda_{FM} = 10$, $\lambda_{prc} = 10$ and $\lambda_{id} = 10$.

\section{Experiments}
The proposed approach is thoroughly validated through a set of experiments aimed at verifying the ability of our model to \textit{(i)} improve the identity preservation when it is applied to the task of reconstructing a face image; \textit{(ii)} hide the identity of another individual while maintaining the face visually unchanged as in an impersonation attack. We employ three different FR models during evaluation: IR152 \cite{deng2019arcface}, MobileFace \cite{chen2018mobilefacenets} and FaceNet \cite{schroff2015facenet}. The pre-trained weights for these models are the same as those used in~\cite{yin2021adv}. 

In order to verify the robustness of our solution to different FR networks, we train two different models: one employs a pre-trained FaceNet model (referred to as Ours-FaceNet) using the implementation and weights from the repository\footnote{\url{https://github.com/timesler/facenet-pytorch}}. The other uses a pre-trained ArcFace model \cite{deng2019arcface} (Ours-ArcFace) taken from the InsightFace repository\footnote{\url{https://github.com/nizhib/pytorch-insightface}}. Note that the pre-trained weights of these models differ from those used for evaluation.

We carry out the experimental validation on the CelebMask-HQ dataset~\cite{lee2020maskgan}, which comprises 30k face images, of which 28k for training, and 2k for testing. Each image is paired with its own semantic segmentation mask, comprising 19 semantic classes.

\subsection{Identity Preservation}\label{subsec:id_preservation}
Injecting the identity embedding in our system has an immediate positive effect: the perceived identity is better preserved during the image generation. This is incidentally a critical issue in generative models for human face generation which often lack this property. To prove that, in Table~\ref{tab:comparison_rec}, we report the average cosine similarity obtained between original and reconstructed faces for different SIS methods on several FR models, which is computed as:

\begin{equation}
    C = \frac{1}{N} \sum_{i=1}^{N} \cos\left(\mathcal{E}_{id}\left(x_i\right), \mathcal{E}_{id}\left(\hat{x_i}\right)\right)
\end{equation}
\noindent
where $\hat{x_i} = \mathcal{G}\left(m_{x_i}, s_{x_i}, id_{x_i}\right)$ is the reconstructed face. A higher cosine similarity score implies an automatic face recognition system will likely verify the two faces as belonging to the same individual. Maintaining higher similarity scores is crucial since face verification systems rely on fixed thresholds to reject or accept face image pairs; the larger the score for genuine pairs, the less the number of false negatives that is returned from the system. 

As expected, state-of-the-art SIS methods struggle to maintain the identity in the generated results. This is especially true for StyleGAN-based methods that, in order to reconstruct and manipulate a given input, rely on GAN-inversion techniques~\cite{zhu2020domain} to find an embedding in the StyleGAN latent space that is as similar as possible to the real image. On the other side, our method exhibits superior performance in identity preservation both w.r.t. to the baseline model $CA^2SIS$~\cite{fontanini2023semantic} and the state of the art. 

In Table~\ref{tab:comparison_rec} we also report the Frechet Inception Distance (FID), which measures the similarity between real and generated data distributions. This is usually employed for validating the quality of the fake samples generated by a model. This is intended to verify that, by conditioning the generation with an identity embedding, we do not compromise the realism of the generated faces. Overall, this conditioning actually negatively impacts on the FID score to a little extent. Although we cannot prove this formally, we believe this is likely due to the way in which FID is calculated: the additional information carried in the identity embedding that is mapped into the pixel space by the generator might shift the fake image distribution, hindering such score. 

\begin{figure}[!t]
    \centering
    \includegraphics[width=\linewidth]{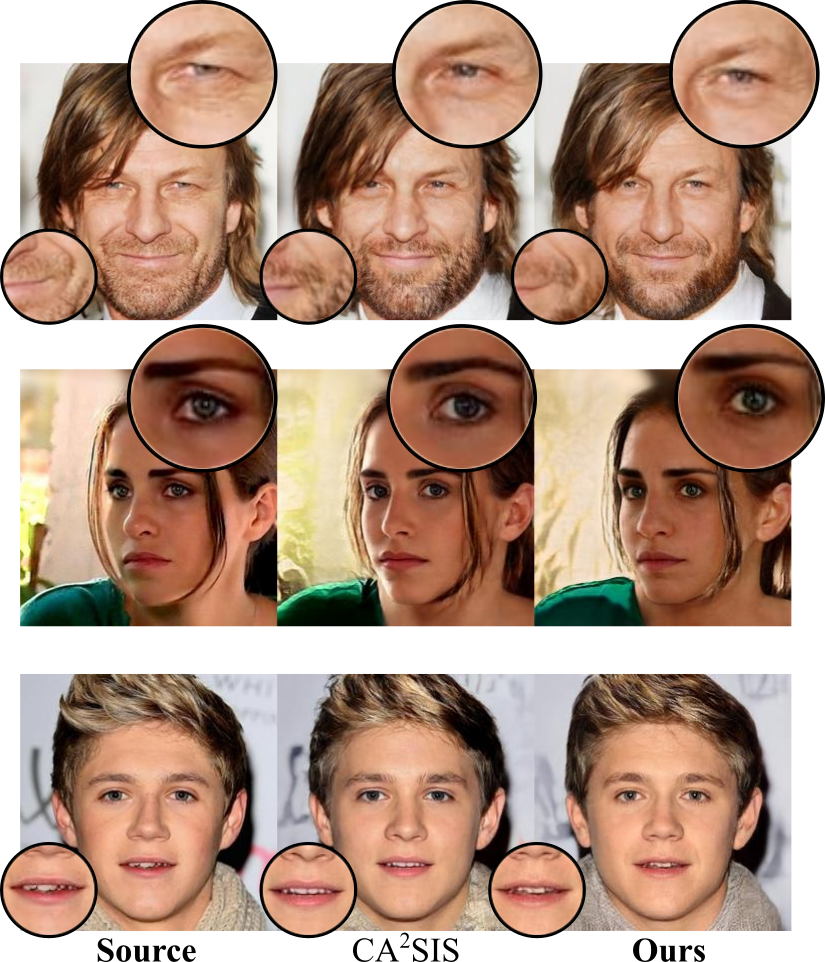}
    \caption{Qualitative comparison between the original $CA^2SIS$ model~\cite{fontanini2023semantic} and the proposed architecture with cross attention-based identity injection (see Sect.~\ref{subsec:id_preservation}). Identity-related details such as the color of the eyes, the eyebrows, and mouth shape or subtle details such as the teeth are better preserved when conditioning with identity embedding. Better seen on screen.}
    \label{fig:id_no_id}
\end{figure}

Nevertheless, even if FID is widely employed to compare different image generation methods, it has also drawn a several criticism \cite{chong2020effectively} and should therefore be always paired with a qualitative evaluation of the results. For this reason, in Fig.~\ref{fig:id_no_id} we report a qualitative example of the effect that the identity injection elicits to our model. At first glance, the results produced by the model with and without identity preservation seem almost identical. Things change if some details are zoomed in and better highlighted. In particular, the eyes and mouth region are the most affected by the identity injection, and more closely resemble the original image. On the other side, the identity information looks almost completely hidden in the generated samples, opening the way to inconspicuous identity swapping as it will be presented in the next section.

\subsection{Adversarial Attack on Face Recognition}\label{sec:attack}
In this section, we show that the proposed architectural change leads to the possibility of generating a face image of some subject $i$ (\textit{attacker}) conditioned with the identity embedding of another individual $j$ (\textit{target}) \textit{i.e.} $\hat{x}_{i\rightarrow j} = \mathcal{G}\left(m_{x_i},s_{x_i},id_{x_j}\right)$.
When doing so, this will be almost completely hidden in the output image, but FR models will recognize the image as belonging to identity $j$. This is very similar to an adversarial attack as in~\cite{jia2022adv}, where the goal is to exploit semantic clues to attack state-of-the-art FR systems. At the same time, there are some key differences: a) our model is not explicitly trained to perform adversarial attacks, but simply for reconstructing an image given its semantic mask, styles and identity; b) it does not require additional training for each attack, but simply swaps the identities of target and attacker at inference time; and, finally, c) the proposed model does not hide the identity into a specific attribute, like eyeglasses as in \cite{jia2022adv}, but treats it as an additional style that is applied globally to the attacker face.

\begin{figure}[!t]
    \centering
    \includegraphics[width=\linewidth]{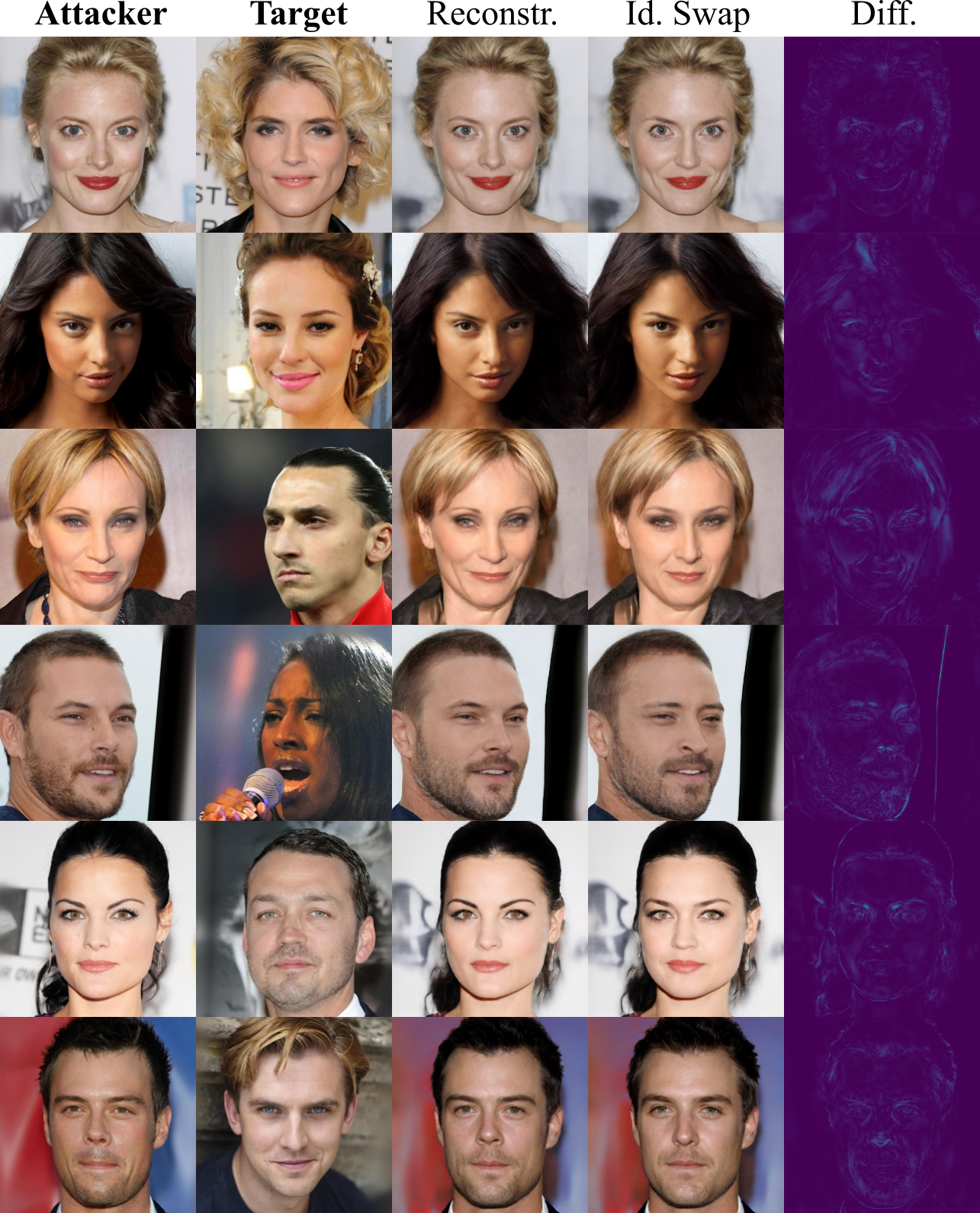}
    \caption{Results of our architecture obtained as described in Sect.~\ref{sec:attack}. The first column is the attacker, the second column is the target, the third column is the reconstruction result of the attacker using the correct identity embedding, fourth column is the reconstruction result when injecting the identity of the target in the attacker. Finally, the last column represents the pixel difference between the two reconstruction results, highlighting that the identity information is effectively concealed in the manipulated face.}
    \label{fig:res}
\end{figure}

Given this task similarity, we evaluate the performance using standard metrics in the field and adopt the Attack Success Rate (ASR) as metric, which is computed as:
\begin{equation}
    \frac{1}{N}\sum_{i=1}^N cos\left(\mathcal{E}_{id}\left(\mathcal{G}\left(m_{x_i},s_{x_i},id_{x_j}\right)\right),\mathcal{E}_{id}\left(x_i\right)\right) > \tau
\end{equation}

\noindent
where $m_{x_i}, s_{x_i}$ are the mask and style codes associated to the attacker $i$, $id_{x_j} = \mathcal{E}_{id}(x_j)$ is the identity embedding associated to the target identity $j$. Finally, $\tau$ value is taken from~\cite{jia2022adv} and is set considering a False Acceptance Rate (FAR) of $0.01$ w.r.t. the attacked FR system. Briefly, this metric quantifies how many times a face recognition systems accepts the input face image as belonging to the target identity and not the attacker.

\begin{figure}[!t]
    \centering
    \includegraphics[width=\linewidth]{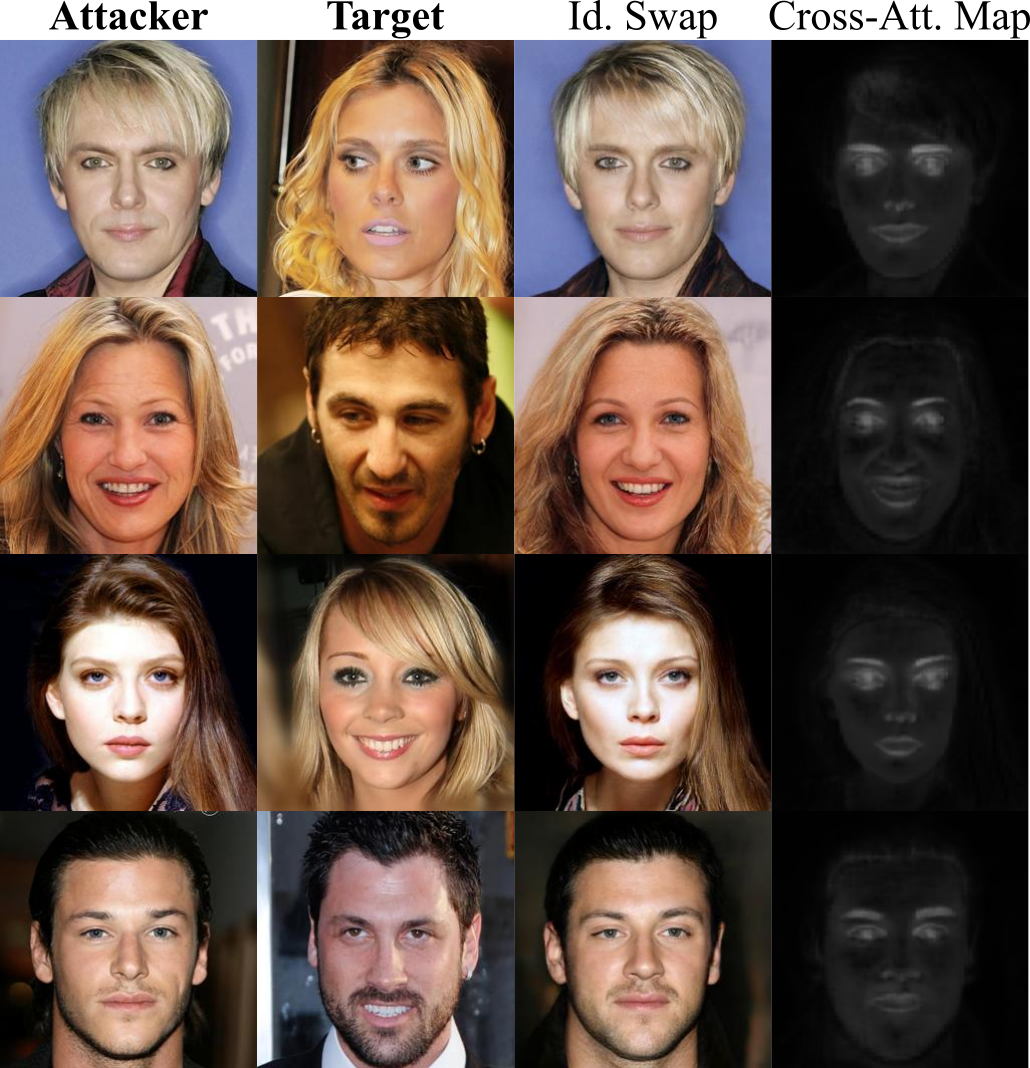}
    \caption{Cross-Attention layer visualization when swapping the identity of a target face to the attacker. The areas that are most affected by the identity injection are the eyes, eyebrows, nose, and mouth. This result suggests the perceived identity is complex information that is carried by several different facial traits.}
    \label{fig:att_map}
\end{figure}

In Fig.~\ref{fig:res} several qualitative results of the adversarial attack are presented: an \textit{attacker} picture is injected with the identity of a \textit{target} face. Indeed, the difference between the attacker face generated with the correct identity and the one generated with the target identity is negligible and a human eye would almost certainly fail when asked to recognize which one was forged with the incorrect identity. The last column of Fig.~\ref{fig:res}, where heatmaps of the pixel-wise $L1$ difference between the two different generated samples are shown, further highlighting this characteristic, exhibiting values close to zero almost everywhere. Still, at a closer look, some minor changes in the swapped samples can be appreciated. In particular, the eyebrows and eyes shape is slightly altered (see third and fourth rows) as well as the nose appearance (see first row). In addition, sometimes also the eye color is affected (see last row). This is in line with the considerations made in~\cite{ferrari2022makes} that the identity information is concentrated in the eyes and eyebrows areas of the face. 

In Fig.~\ref{fig:att_map}, we show the cross-attention maps estimated by the model when injecting identity information, in order to visualize in which face areas the identity gets mapped by the attention layer. This figure suggests our intuition was correct, that is a high-level complex information such as the identity influences several different regions of the face, making state-of-the-art SIS methods based on SPADE unsuitable for this task.

Finally, in Table~\ref{tab:comparison} an extensive comparison with state-of-the-art methods for adversarial attack on human face recognition is presented. All the numbers are taken from~\cite{jia2022adv}, and our results are calculated using the same settings and pre-trained FR models. Our method is able to improve the ASR score of the state-of-the-art, by a great margin in some cases. When the attacked FR architecture is the same as that used to train our model, the improvement is clearly larger. We note this is a fair setting since the architecture is the same, but the pre-trained weights are not. Thus, it resembles a \textit{transferable adversarial attack} (or gray-box) scenario, that is when the attacker is aware of the victim's architecture but does not have access to its internal weights. In the other cases (the attacked FR model is different from that of our architecture), our solution still obtains largely higher ASR score in all the cases except one. Other than demonstrating that our solution is highly effective, it suggests that a significant overlap across recognition models to the relevant facial features useful for recognition occurs. Compared to the state-of-the-art, our results are quite impressive considering that we do not specifically train our system to perform adversarial attacks (nor it requires fine-tuning). Differently, all the compared methods perform a specific optimization.

\begin{table}[!t]
    \centering
    \begin{adjustbox}{width=\linewidth}
    \begin{tabular}{c|c|c|c}
    \toprule
        \textbf{Method} & \textbf{IR152} $\uparrow$ & \textbf{MobileFace}  $\uparrow$ & \textbf{FaceNet}  $\uparrow$ \\
        \midrule
        FGSM \cite{goodfellow2014explaining} & 2.70 & 5.10 & 1.90 \\
        
        PGD \cite{madry2017towards} & 26.00 & 29.90 & 3.50 \\
        
        MI-FGSM \cite{dong2018boosting} & 26.80 & 21.70 & 4.60 \\
        
        C\&W \cite{carlini2017towards} &  27.30 & 28.20 & 3.30 \\
        \midrule
        Adv-Hat \cite{komkov2021advhat} & 2.50 & 8.40 & 4.70 \\
        
        Adv-Glasses \cite{sharif2016accessorize} & 4.50 & 5.60 & 9.10 \\
        
        Gen-AP \cite{xiao2021improving} &  19.50 & 24.40 & 15.80 \\
        \midrule
        Adv-Face \cite{deb2020advfaces} & 31.40 & 36.40 & 21.60 \\
        
        Adv-Makeup \cite{yin2021adv} & 10.80 & 14.60 & 10.50 \\
        
        Semantic-Adv \cite{qiu2020semanticadv} & 10.30 & 19.40 & 9.00 \\
        
        Adv-Attribute \cite{jia2022adv} & 44.30 & \underline{50.20} & 31.80 \\
        \midrule
        \midrule
        \textbf{Ours} - FaceNet & \underline{48.30} & 40.60 & \textit{\textbf{77.60}} \\
        \textbf{Ours} - ArcFace & \textit{\textbf{67.80}} & \textbf{98.10} & \underline{72.20} \\

        \bottomrule
    \end{tabular}
    \end{adjustbox}
    \caption{ASR comparisons against adversarial attacks methods targeting different models on CelebA-HQ, as detailed in Sect.~\ref{sec:attack}. \textbf{Bold}=best result, \underline{underlined}=second best. \textit{Italic} font indicates the attacked FR architecture is the same as that used to train our model, but the pre-trained weights differ.}
    \label{tab:comparison}
\end{table}

\begin{figure}[bh!]
    \centering
    \includegraphics[width=\linewidth]{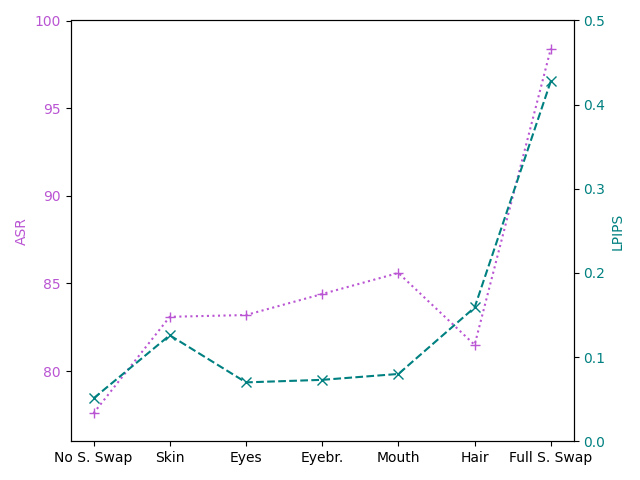}
    \caption{Graph showing different Attack Success Rate (ASR) and LPIPS metric values when swapping different styles along the identity. The style swapping procedure is described in Sect.~\ref{sec:swap_styles}.}
    \label{fig:style_graph}
\end{figure}

\subsection{Style Transfer Effect on Face Recognition} \label{sec:swap_styles}
The SIS architecture, which was adapted to include identity information during generation, extracts a set of styles $s_{x_i}$ via $\mathcal{E}_s$ from each semantic region of an RGB image and maps them to the corresponding semantic class in the input mask (see Sect.~\ref{sec:ID_SIS}). This allows the style transfer between a source $x_i$ and a target $x_j$ image by mixing their corresponding style codes $s_{x_i}$ and $s_{x_j}$ \textit{e.g.} hair color. 

\begin{figure*}[!t]
    \centering
    \includegraphics[width=\linewidth]{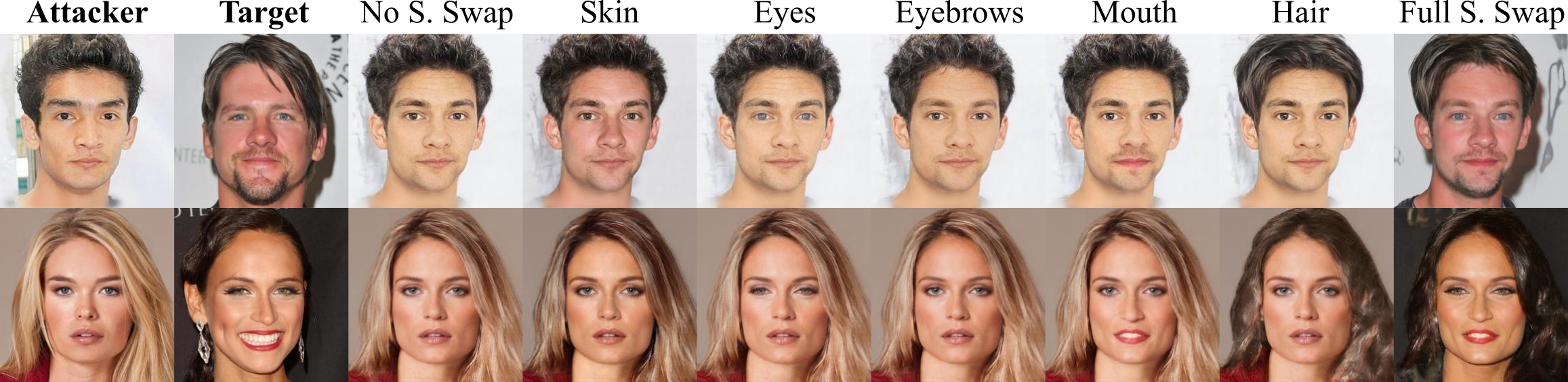}
    \caption{Results of different types of style swaps as described in Sect.~\ref{sec:swap_styles}.}
    \label{fig:style_swap}
\end{figure*}

In this section, we explore how style transfer affects the attacks on FR. More in detail, the objective is to verify if swapping some styles can boost the ASR. To prove this, firstly, we performed style transfer on a set of styles combined with identity swap. In particular, \textit{Skin}, \textit{Eyes}, \textit{Eyebrows}, \textit{Mouth}, \textit{Hair} and \textit{Full S. Swap} (\textit{i.e.} all the styles are swapped) were chosen.  Then, the results for each different style transfer experiment were evaluated using two different metrics: ASR and LPIPS. More in detail, the LPIPS metric was calculated between the images reconstructed with the correct identity and the ones generated with style transfer and identity swap. By doing so, we can measure how much the style transfer alters the overall appearance of the generated samples. Ideally, the best results are those in which the system is able to fool the FR models while maintaining the attack almost invisible to the human eye; in other words, we want the ASR to be as high as possible, while the LPIPS remains as low as possible

Quantitative results of this experiment can be seen in Fig.~\ref{fig:style_graph}, while qualitative results can be seen in Fig.~\ref{fig:style_swap}.
In both figures, the ``\textit{No S. Swap}" value represents the baseline, \textit{i.e.} the identity swapping is performed without any style transfer. 
As expected in this case, the LPIPS value is very low enforcing our claim of inconspicuous attacks. On the other side, when performing style transfer, ASR results increase for every different transferred part. This proves that the capability of pairing identity and style swaps can strengthen the attacks against FR systems giving our model an additional edge w.r.t. current state-of-the-art.

Interestingly, the parts that increased the ASR results the most are also the ones with the lower LPIPS. Indeed, they are also the ones that contain the most identity information. More in detail, the single semantic classes that obtained the highest ASR results were \textit{Eyes}, \textit{Eyebrows} and \textit{Mouth}. This is in line with the results presented in Fig.~\ref{fig:id_no_id}. These findings prove that is possible to combine style and identity swaps maintaining the attack almost invisible to the human eye, which is of paramount importance for this kind of system.
Finally, when swapping all the styles together (``\textit{Full S. Swap}" in the figure), the ASR reaches $98.4$\% but, at the same time, LPIPS value is the highest.

\section{Conclusions and Ethical Concerns}
In this paper, we proposed a novel Semantic Image Synthesis (SIS) method for image manipulation that also employs identity information during the generation process. 
The identity injection into the model is based on the identity embedding, extracted from a pre-trained FR system, as an additional style that is concatenated to the other styles obtained by a style encoder from a reference RGB image. Then, a cross-attention mechanism is used in the generator.
We observe the proposed identity injection procedure has two main effects: firstly, when the same identity of the input subject is used, it greatly improves the identity preservation during generation. 
Secondly, if the identity is swapped (\ie the injected identity is different with respect to the input one), the model is able to perform an adversarial attack to FR systems hindering their results. Extensive experiments on these two contributions were performed proving the effectiveness of the proposed architecture. 

As a future work, we plan to further develop the identity injection mechanism, so as to have a strategy for making the identity of the second subject in the final generated image visible or not. This is important both for controlling the extent of the identity injected into the system and for the development and study of biometric systems based on facial identification, such as face swapping and morphing. 

Lastly, we are aware that these types of systems could be used in malicious or criminal ways. On the other side, we strongly believe that studying and publicly sharing results in this field can increase awareness of the use of these systems in the academic community (and beyond), stimulate the development of new countermeasures, and lead to the creation of new datasets for training future systems.

\section*{Acknowledgments}
This work was partially supported by PRIN 2020 “LEGO.AI: LEarning the Geometry of knOwledge in AI systems”, grant no. 2020TA3K9N funded by the Italian MIUR. Additionally, this work was partially supported by ``Partenariato FAIR (Future Artificial 
Intelligence Research) - PE00000013, CUP J33C22002830006" funded by the European Union - NextGenerationEU through the italian MUR within NRRP.

{
    \small
    \bibliographystyle{ieeenat_fullname}
    \bibliography{main}
}

\end{document}